\documentclass[sigconf,nonacm]{acmart}
\usepackage{makecell}
\usepackage{microtype}
\AtBeginDocument{%
  }

\setcopyright{acmlicensed}
\copyrightyear{2018}
\acmYear{2018}
\acmDOI{XXXXXXX.XXXXXXX}
\acmConference[Conference acronym 'XX]{Make sure to enter the correct
  conference title from your rights confirmation email}{June 03--05,
  2018}{Woodstock, NY}
\acmISBN{978-1-4503-XXXX-X/2018/06}

\begin{document}

\title{SciFigDetect: A Benchmark for AI-Generated Scientific Figure Detection}

\author{You Hu}
\authornote{Both authors contributed equally to this research.}
\affiliation{%
  \institution{Zhejiang University}
  \country{Hangzhou, China}}
\email{huyou@zju.edu.cn}

\author{Chenzhuo Zhao}
\authornotemark[1]
\affiliation{%
  \institution{Independent Researcher}
  \country{Beijing, China}
}
\email{cyczzhao@gmail.com}

\author{Changfa Mo}
\affiliation{%
  \institution{Zhejiang University}
  \country{Hangzhou, China}}
\email{monge_zju@zju.edu.cn}

\author{Haotian Liu}
\affiliation{%
 \institution{University of Oulu}
 \country{Oulu, FINLAND}}
\email{Haotian.Liu@oulu.fi}

\author{Xiaobai Li}
\authornote{Corresponding author.}
\affiliation{%
  \institution{Zhejiang University}
  \country{Hangzhou, China}
  }
\email{xiaobai.li@zju.edu.cn}

\begin{abstract}
Modern multimodal generators can now produce scientific figures at near-publishable quality, creating a new challenge for visual forensics and research integrity. Unlike conventional AI-generated natural images, scientific figures are structured, text-dense, and tightly aligned with scholarly semantics, making them a distinct and difficult detection target. However, existing AI-generated image detection benchmarks and methods are almost entirely developed for open-domain imagery, leaving this setting largely unexplored.
We present the first benchmark for AI-generated scientific figure detection. To construct it, we develop an agent-based data pipeline that retrieves licensed source papers, performs multimodal understanding of paper text and figures, builds structured prompts, synthesizes candidate figures, and filters them through a review-driven refinement loop. The resulting benchmark covers multiple figure categories, multiple generation sources and aligned real--synthetic pairs.
We benchmark representative detectors under zero-shot, cross-generator, and degraded-image settings. Results show that current methods fail dramatically in zero-shot transfer, exhibit strong generator-specific overfitting, and remain fragile under common post-processing corruptions. These findings reveal a substantial gap between existing AIGI detection capabilities and the emerging distribution of high-quality scientific figures. We hope this benchmark can serve as a foundation for future research on robust and generalizable scientific-figure forensics. The dataset is available at https://github.com/Joyce-yoyo/SciFigDetect.
\end{abstract}  


\begin{teaserfigure}
  \includegraphics[width=0.95\textwidth]{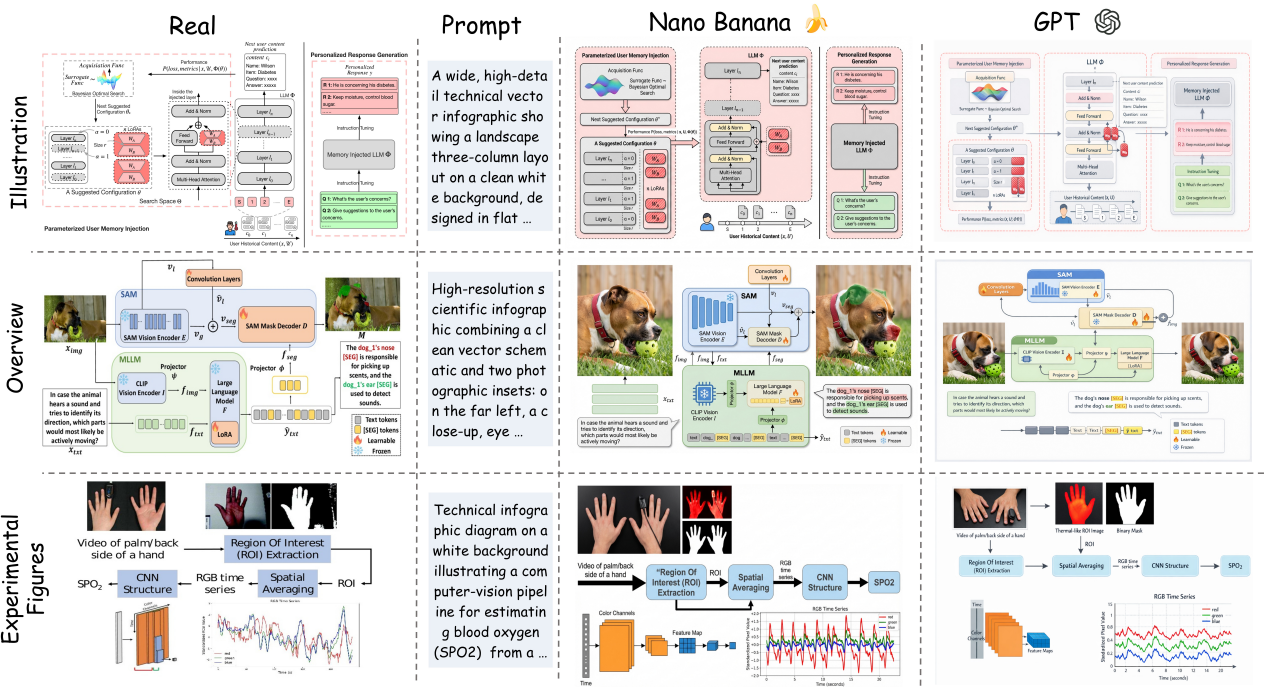}
  \caption{\textbf{Overview of our benchmark.} Representative real--synthetic examples from three figure categories: \emph{Illustration}, \emph{Overview}, and \emph{Experimental Figures}. For each case, we show the real figure, the prompt, and synthetic counterparts generated by \textbf{Nano Banana} and \textbf{GPT}.}
  \label{fig:teaser}
\end{teaserfigure}


\maketitle

\begin{table*}[t]
\centering
\caption{Comparison with representative AI-generated image detection datasets. Our benchmark differs from prior work by focusing on structured, text-dense scientific figures and by explicitly preserving figure-related paper context.}
\label{tab:dataset_comparison}
\resizebox{\textwidth}{!}{
\begin{tabular}{l c c c c c c c c}
\toprule
Dataset
& \makecell{Image\\Content}
& \makecell{Generator\\Source}
& \makecell{Public\\Availability}
& \makecell{Real\\Images}
& \makecell{Fake\\Images}
& Resolution
& \makecell{Structured /\\Text-Dense}
& \makecell{Paper-Context\\Aware} \\
\midrule
UADFV~\cite{yang2018exposingdeepfakesusing} & Face & GAN & $\times$ & 241 & 252 & Varied & $\times$ & $\times$ \\
FakeSpotter~\cite{wang2020fakespottersimplerobustbaseline} & Face & GAN & $\times$ & 6,000 & 5,000 & Varied & $\times$ & $\times$ \\
DFFD~\cite{dang2020detectiondigitalfacemanipulation} & Face & GAN & $\checkmark$ & 58,703 & 240,336 & Varied & $\times$ & $\times$ \\
APFDD~\cite{gandhi2020adversarialperturbationsfooldeepfake} & Face & GAN & $\times$ & 5,000 & 5,000 & Varied & $\times$ & $\times$ \\
ForgeryNet~\cite{he2021forgerynetversatilebenchmarkcomprehensive} & Face & GAN & $\checkmark$ & 1,438,201 & 1,457,861 & Varied & $\times$ & $\times$ \\
DeepArt~\cite{wang2023benchmarkingdeepartdetection} & Art & Diffusion  & $\checkmark$ & 64,479 & 73,411 & Varied & $\times$ & $\times$ \\
CNNSpot~\cite{wang2019cnngenerated} & General & ProGAN (GAN) & $\checkmark$ & 362,000 & 362,000 & Varied & $\times$ & $\times$ \\
DE-FAKE~\cite{sha2023defakedetectionattributionfake} & General & Diffusion & $\times$ & 20,000 & 60,000 & Varied & $\times$ & $\times$ \\
CIFAKE~\cite{bird2023cifakeimageclassificationexplainable} & General & Stable Diffusion v1.4 & $\checkmark$ & 60,000 & 60,000 & 32$\times$32 & $\times$ & $\times$ \\
GenImage~\cite{zhu2023genimage} & General & GAN + Diffusion & $\checkmark$ & 1,331,167 & 1,350,000 & 512$\times$512 & $\times$ & $\times$ \\
\midrule
\textbf{Ours} & \textbf{Scientific figures} & \makecell{\textbf{Nano Banana Pro} \\ \textbf{GPT-image-1.5}} & \textbf{\checkmark} & \textbf{72,965} & \textbf{150,807} & \textbf{Aligned} & \textbf{\checkmark} & \textbf{\checkmark} \\
\bottomrule
\end{tabular}
}
\end{table*}

\section{Introduction}
\label{sec:intro}

Recent advances in generative models enable the synthesis of scientific figures at near-publishable  quality~\cite{zhu2026autofiguregeneratingrefiningpublicationready}. Modern multimodal systems no longer generate only generic synthetic imagery; they can now produce academic illustrations that exhibit high visual quality, structural coherence, and semantic consistency with scientific narratives. In particular, systems such as Nano Banana~\cite{google2025nanobananapro} and ChatGPT-like multimodal generators~\cite{openai2022chatgpt,openai2025gpt5,openai_gptimage1_docs} support iterative generation and refinement workflows that make synthetic scientific figures increasingly realistic and usable in academic writing~\cite{zhu2026paperbananaautomatingacademicillustration,zhu2026autofiguregeneratingrefiningpublicationready}.


This is already becoming a real integrity risk. Major publishers and venues have begun to restrict or prohibit AI-generated figures in submissions and publications~\cite{nature_ai_policy,cell_figure_guidelines,science_editorial_policies,eurognc_ai_policy,europar2026cfp}. As generation quality improves, these systems can be used to fabricate visual evidence, insert generated figures without disclosure, or reproduce scientific visuals without attribution. Because figures often communicate methods, results, and evidence in condensed form, such misuse can directly undermine trust in scientific communication.

Detecting such images poses distinct challenges compared to conventional AI-generated image detection. Scientific figures differ from  natural images. They are structured, text-dense, symbol-heavy, and tightly coupled with scholarly semantics~\cite{hsu2021scicapgeneratingcaptionsscientific,li2024multimodalarxivdatasetimproving,roberts2024scifibenchbenchmarkinglargemultimodal,li2025mmscidatasetgraduatelevelmultidiscipline,zhao2025multimodalfoundationmodelsunderstand}. Their visual structure is governed less by natural image statistics than by established conventions for scientific communication. As a result, detectors developed for faces, scenes, and generic AIGC content~\cite{wang2019cnngenerated,ojha2023towards,tan2023learning,tan2024rethinking,tan2024frequency,liu2024forgery,yan2024sanity,yan2024effort} may not transfer to this setting. The cues they rely on, such as texture artifacts, frequency irregularities, or stylistic inconsistencies~\cite{durall2020watch,jeong2022bihpf,tan2023learning,tan2024rethinking},  are often weak or unstable in high-quality academic figures.  The problem is further compounded by the fact that generated scientific figures intentionally mimic the layouts, annotation density, and semantic structure of real ones, making the boundary between authentic and generated scientific figures more subtle than in standard AIGC benchmarks.

However, current AI-generated image detection benchmarks are poorly matched to this threat model, which focus on open-domain content such as portraits, natural scenes, and generic objects~\cite{wang2019cnngenerated,ojha2023towards,yan2024effort}, and therefore fail to capture the distinctive properties of scientific figures, including high text density, strong structural constraints, semantic precision, and publication-oriented refinement. Consequently, strong performance on existing benchmarks does not imply robustness in academic settings.

In this paper, we introduce \textbf{SciFigDetect}, the first benchmark for AI-generated scientific figure detection. Our benchmark targets high-quality scientific figures generated under realistic academic workflows and is constructed from commercially permissible open-access papers through an \textbf{agent-based data pipeline}. Starting from licensed source papers, the pipeline performs multimodal understanding of paper text and figures, derives structure-aware prompts, synthesizes candidate figures with modern generators, and filters them through a review-driven refinement loop. The resulting benchmark preserves figure-related paper context and provenance metadata, and covers multiple figure categories, multiple generation sources, and aligned real--synthetic pairs. In total, it contains 72,965 real figures and 150,807 synthetic figures across three representative categories: \textbf{Illustration}, \textbf{Overview}, and \textbf{Experimental Figure}.


We use SciFigDetect to evaluate representative detectors under \textbf{zero-shot}, \textbf{cross-generator}, and \textbf{degraded-image} settings. The results are clear. Existing detectors fail dramatically in the zero-shot setting. They overfit strongly to seen generators and transfer poorly across Nano Banana and GPT-generated figures. Their performance also drops substantially under common post-processing degradations such as compression, blur, and noise. These results expose a large gap between existing AIGI detection methods and the emerging distribution of high-quality scientific figures, and suggest that scientific figure forensics remains largely unsolved.

In summary, our contributions are three-fold:
\begin{itemize}
    \item We introduce SciFigDetect, the first benchmark for AI-generated scientific figure detection, featuring high-quality scientific figures generated under realistic academic workflows.
    \item We develop a scalable and compliant agent-based data construction pipeline for constructing realistic scientific figure benchmarks.
    \item We provide a comprehensive evaluation of representative detectors under zero-shot, cross-generator, and degraded-image settings, revealing substantial generalization and robustness limitations.
\end{itemize}

\label{sec:intro}

\section{Related Work}
\label{sec:related_work}

\subsection{AI-generated scientific illustrations}
Recent work has begun to explore the automatic generation of publication-ready scientific figures using multimodal models and agentic pipelines~\cite{mondal2024scidoc2diagrammermafgenerationscientificdiagrams,wei2024wordsstructuredvisualsbenchmark,lin2026autofigureeditgeneratingeditablescientific,huang2026scifigautomatingscientificfigure}. PaperBanana presents an agentic framework for academic illustration generation, where specialized agents retrieve references, plan content and style, render images, and iteratively refine outputs through self-critique~\cite{zhu2026paperbananaautomatingacademicillustration}. AutoFigure further advances this direction by introducing a benchmark of scientific text--figure pairs together with an agentic framework for generating and refining scientific illustrations from long-form scientific text~\cite{zhu2026autofiguregeneratingrefiningpublicationready}. These works suggest that high-quality academic figure synthesis is becoming practical. However, their primary focus is on generation quality rather than detection robustness.

\subsection{AI-generated image detection}
AI-generated image detection has been studied extensively in open-domain settings such as faces, natural scenes, artworks, and generic synthetic imagery~\cite{dang2020detectiondigitalfacemanipulation,wang2020fakespottersimplerobustbaseline}. As shown in Table~\ref{tab:dataset_comparison}, the field has evolved from early benchmarks dominated by GAN-generated images~\cite{yang2018exposingdeepfakesusing,dang2020detectiondigitalfacemanipulation,goodfellow2014generative,karras2018progressive,karras2019style,karras2020analyzing} to more recent datasets covering diffusion-based~\cite{dhariwal2021diffusion,gu2022vector,ho2020denoising,nichol2021glide} and text-to-image generation. Early datasets were often limited in scale or domain coverage, while later benchmarks such as CIFAKE, DeepArt, DE-FAKE, and GenImage expanded both generator diversity and image distribution complexity~\cite{bird2023cifakeimageclassificationexplainable,wang2023benchmarkingdeepartdetection,sha2023defakedetectionattributionfake,zhu2023genimage}. GenImage is particularly notable for introducing large-scale evaluation together with cross-generator and degraded image protocols~\cite{zhu2023genimage}.
Methodologically, existing detectors mainly exploit spatial artifacts~\cite{wang2019cnngenerated,chai2020makes}, frequency-domain irregularities~\cite{durall2020watch,jeong2022bihpf,tan2024frequency}, gradient cues~\cite{tan2023learning}, structural artifacts such as neighboring pixel relationships~\cite{tan2024rethinking}, or CLIP-based foundation model representations~\cite{radford2021learning,ojha2023towards,liu2024forgery,yan2024sanity,yan2024effort}. However, both existing datasets and detectors are largely designed for open-domain synthetic images. It remains unclear whether they transfer to scientific figures, which are text-dense, structurally constrained, and tightly aligned with scholarly semantics. To the best of our knowledge, prior work has not systematically studied the detection of high-quality AI-generated scientific figures produced by systems such as Nano Banana and ChatGPT-like multimodal generators. Our work fills this gap by introducing SciFigDetect dedicated to AI-generated scientific figure detection and by evaluating existing detectors under zero-shot, cross-generator, and robustness settings.

\section{Dataset Construction}
\label{sec:method}


\subsection{Problem Setup}

Our goal is to construct a realistic benchmark for detecting AI-generated scientific figures. Starting from open-access academic papers, we build a compliant pipeline that pairs real scientific figures with synthetic counterparts generated by modern image-generation models, while preserving figure-related context and provenance metadata.

Formally let
$
\mathcal{P}=\{p_i\}_{i=1}^{N}
$
denote the candidate paper pool, where each paper $p_i$ contains the PDF, metadata, and license information. We retain only papers released under commercially permissible licenses:
\begin{equation}
\mathcal{P}^{+}=\{p_i \in \mathcal{P}\mid \ell_i \in \mathcal{L}_{\mathrm{perm}}\},
\label{eq:license_filter}
\end{equation}
where $\ell_i$ is the license of paper $p_i$, and $\mathcal{L}_{\mathrm{perm}}$ denotes the set of permitted licenses, such as CC BY.

For each retained paper $p\in\mathcal{P}^{+}$, we construct a benchmark sample
\begin{equation}
z=\big(c,\; f_{\mathrm{real}},\; f_{\mathrm{syn}},\; a\big),
\label{eq:sample_def}
\end{equation}
where $c$ denotes the figure-related paper context, $f_{\mathrm{real}}$ the original figure, $f_{\mathrm{syn}}$ the accepted synthetic figure, and $a$ auxiliary metadata. The final benchmark is
\begin{equation}
\mathcal{D}_{\mathrm{bench}}=\{z_n\}_{n=1}^{|\mathcal{D}_{\mathrm{bench}}|}.
\end{equation}

\begin{figure*}[t!]
\centering
\includegraphics[width=0.95\linewidth]{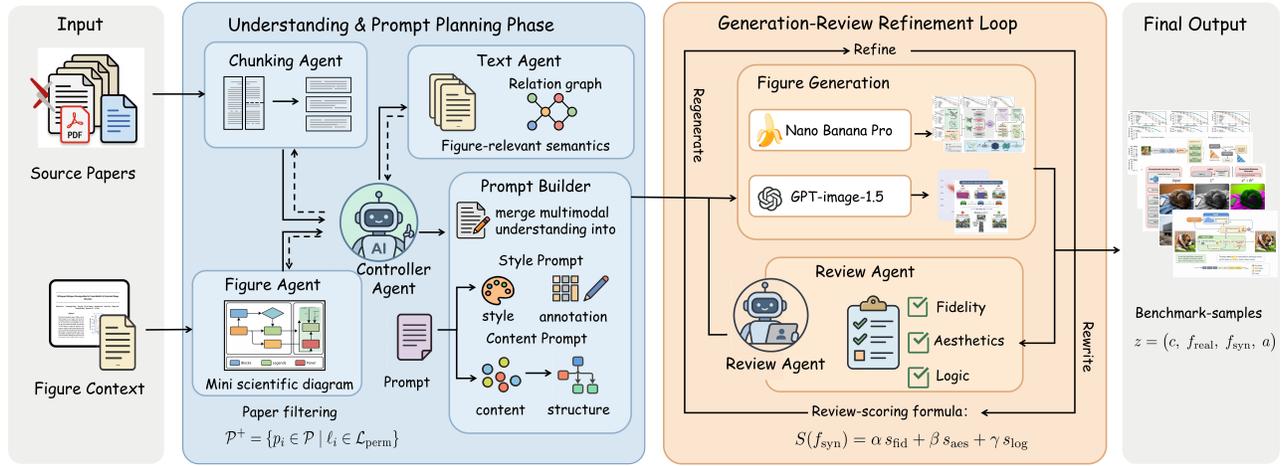}
\caption{\textbf{Overview of the data construction pipeline.}
From licensed source papers and figure-related context, our framework performs multimodal understanding, prompt planning, and iterative generation--review refinement to construct benchmark samples $z=(c, f_{\mathrm{real}}, f_{\mathrm{syn}}, a)$.}
\label{fig:architecture}
\end{figure*}

\subsection{Construction Pipeline}

Figure~\ref{fig:architecture} shows the overall pipeline, which consists of two stages: an \emph{Understanding \& Prompt Planning Phase} and a \emph{Generation--Review Refinement Loop}. The framework adopts a master--worker architecture, where a GPT-based Controller Agent coordinates specialized worker agents for chunking, text understanding, figure understanding, prompt construction, generation, and review.

\paragraph{Understanding \& Prompt Planning.}
Given a source paper and its figure-related context, the Chunking Agent first segments the paper into semantically coherent chunks according to section boundaries, paragraph continuity, and figure-reference relations. The Text Agent then extracts figure-relevant semantics from the paper body, including research background, workflow, entities, and structural relations. In parallel, the Figure Agent analyzes the original scientific figure itself, focusing on layout composition, module organization, arrows, legends, color usage, and spatial hierarchy, and also assigns a coarse figure-type label such as illustration, overview, or experimental figure. The Prompt Builder merges these multimodal signals into a structured prompt with style-oriented and content-oriented components, capturing visual conventions, annotation style, semantic entities, and scientific structure.

\paragraph{Generation--Review Refinement Loop.}
Given the structured prompt, the Figure Generation module synthesizes candidate scientific figures using models such as Nano Banana Pro and GPT-image-1.5. The goal is not pixel-level reconstruction, but the generation of new figures that preserve the core semantics, logical organization, and plausible academic style of the source content. Generated candidates are then scored by a Review Agent from three aspects: academic fidelity, aesthetic consistency, and logical coherence. We define the overall review score as
\begin{equation}
S(f_{\mathrm{syn}})=\alpha\, s_{\mathrm{fid}}+\beta\, s_{\mathrm{aes}}+\gamma\, s_{\mathrm{log}},
\label{eq:review_score}
\end{equation}

where $s_{\mathrm{fid}}$, $s_{\mathrm{aes}}$, and $s_{\mathrm{log}}$ denote the three review scores, and $\alpha+\beta+\gamma=1$. In our implementation, we set $\alpha=\beta=\gamma=\frac{1}{3}$, assigning equal importance to the three criteria. A candidate is accepted only if its overall review score is at least $0.6$. Candidates that fail review are either \emph{rewritten} by revising the prompt or \emph{regenerated} by re-sampling from the generator under the Controller's guidance, forming a closed-loop refinement process~\cite{madaan2023selfrefineiterativerefinementselffeedback}.

\paragraph{Dataset Curation.}
Once a candidate passes review, it is stored as a benchmark sample following Eq.~\eqref{eq:sample_def}. Each sample includes the paper context, the original figure, the accepted synthetic figure, and auxiliary metadata such as figure category, prompt, license information, generator identity, and review history.

\begin{figure}[h]
\centering
\includegraphics[width=\columnwidth]{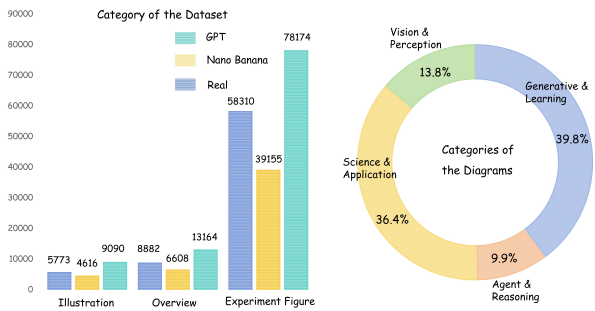}
\caption{\textbf{Dataset statistics.} Left: per-category sample counts for real, Nano Banana, and GPT-generated figures. Right: topic distribution of the collected diagrams. Together, they show both the scale and diversity of our benchmark.}
\label{fig:category_dist}
\end{figure}

\subsection{Dataset Statistics}

Figure~\ref{fig:category_dist} summarizes the composition of our benchmark from two perspectives. The left panel reports the sample counts of real, Nano Banana, and GPT-generated figures across three figure types: Illustration, Overview, and Experimental Figure. Specifically, the real subset contains 5,773 illustrations, 8,882 overviews, and 58,310 experimental figures; the Nano Banana subset contains 4,616 illustrations, 6,608 overviews, and 39,155 experimental figures; and the GPT subset contains 9,090 illustrations, 13,164 overviews, and 78,174 experimental figures.

A key subset of the benchmark forms aligned true--synth pairs, where the same source figure is associated with both Nano Banana and GPT-generated counterparts. The numbers of such paired samples are 4,616 for illustrations, 6,608 for overviews, and 39,155 for experimental figures. These aligned pairs enable controlled comparison between real figures and multiple synthetic variants derived from the same paper context.

The right panel of Fig.~\ref{fig:category_dist} shows the topical distribution of the collected diagrams. The dataset spans four major groups: Generative \& Learning (7,079, 39.8\%), Science \& Application (6,478, 36.4\%), Vision \& Perception (2,459, 13.8\%), and Agent \& Reasoning (1,769, 9.9\%). This distribution indicates that the benchmark covers diverse scientific content rather than a single narrow topic.

Overall, our benchmark combines multiple figure types, multiple generation sources, aligned true--synth pairs, and broad topical coverage, providing a realistic and diverse testbed for scientific-figure detection. Additional implementation details are provided in the supplementary material.

\begin{table*}[h]
\centering
\caption{Zero-shot performance of existing AI-generated image detectors on SciFigDetect. 
Models are evaluated without any adaptation to scientific figures.
$\mathrm{Acc}_{real}$ denotes accuracy on real figures, and $\mathrm{Acc}_{fake}$ summarizes accuracy on synthetic figures. 
$\mathrm{AvgAcc}$ is the unweighted mean of $\mathrm{Acc}_{real}$, $\mathrm{Acc}_{GPT}$, and $\mathrm{Acc}_{Banana}$. 
 Best and second-best results are highlighted in bold and underline, respectively.}
\label{tab:zeroshot_main}
\resizebox{\textwidth}{!}{
\begin{tabular}{lcccc|cccc|cccc|cccc}
\toprule
& \multicolumn{4}{c|}{All Figures}
& \multicolumn{4}{c|}{Illustrations}
& \multicolumn{4}{c|}{Overviews}
& \multicolumn{4}{c}{Experimental Figures} \\
\cmidrule(lr){2-5} \cmidrule(lr){6-9} \cmidrule(lr){10-13} \cmidrule(lr){14-17}
Method
& Acc$_\text{real}$ & Acc$_\text{fake}$ & $\mathrm{AvgAcc}$ & AP
& Acc$_\text{real}$ & Acc$_\text{fake}$ & $\mathrm{AvgAcc}$ & AP
& Acc$_\text{real}$ & Acc$_\text{fake}$ & $\mathrm{AvgAcc}$ & AP
& Acc$_\text{real}$ & Acc$_\text{fake}$ & $\mathrm{AvgAcc}$ & AP \\
\midrule
CNNSpot~\cite{wang2019cnngenerated}
& 99.60 & 2.74 & 35.03 & 72.19
& \textbf{100.00} & 1.20 & 34.13 & 72.12
& \textbf{100.00} & 1.22 & 34.14 & 74.80
& 99.49 & 3.17 & 35.28 & 72.23 \\
PatchFor~\cite{chai2020makes}
& \underline{99.94} & 0.16 & 33.42 & 67.29
& \textbf{100.00} & 0.11 & 33.41 & 67.40
& \underline{99.85} & 0.08 & 33.33 & 66.47
& \underline{99.95} & 0.18 & 33.44 & 67.61 \\
UniFD~\cite{ojha2023towards}
& 89.00 & 7.80 & \underline{48.40} & 49.11
& 96.73 & 2.07 & \underline{49.40} & 49.15
& 97.57 & 1.90 & \underline{49.73} & 46.41
& 86.19 & 8.39 & \underline{47.29} & 45.40 \\
LGrad~\cite{tan2023learning}
& 98.89 & 8.48 & \textbf{53.68} & 65.30
& \underline{99.78} & 8.17 & \textbf{53.98} & 75.40
& 99.39 & 11.17 & \textbf{55.28} & \underline{80.87}
& 98.72 & 8.09 & \textbf{53.40} & 61.81 \\
NPR~\cite{tan2024rethinking}
& 94.92 & \underline{13.24} & 40.47 & \underline{75.33}
& 98.26 & \underline{11.76} & 40.60 & \textbf{80.27}
& 99.24 & \underline{13.07} & 41.79 & \textbf{84.34}
& 93.81 & \underline{13.44} & 40.23 & \underline{73.76} \\
FreqNet~\cite{tan2024frequency}
& 98.19 & 2.76 & 34.57 & 72.06
& 99.13 & 1.42 & 33.99 & 76.43
& \textbf{100.00} & 1.37 & 34.25 & 79.88
& 97.77 & 3.15 & 34.69 & 70.53 \\
FatFormer~\cite{liu2024forgery}
& 96.42 & 1.25 & 32.98 & 60.86
& 96.30 & 0.65 & 32.53 & 61.75
& 98.48 & 0.61 & 33.23 & 60.97
& 96.08 & 1.43 & 32.98 & 60.62 \\
AIDE~\cite{yan2024sanity}
& 88.69 & \textbf{15.26} & 39.74 & 68.67
& 81.26 & \textbf{17.10} & 38.49 & 67.67
& 86.47 & \textbf{19.07} & 41.54 & 69.01
& 89.94 & \textbf{14.40} & 39.58 & 68.78 \\
Effort~\cite{yan2024effort}
& \textbf{100.00} & 0.95 & 33.96 & \textbf{76.48}
& \textbf{100.00} & 0.00 & 33.33 & \underline{77.45}
& \textbf{100.00} & 0.00 & 33.33 & 75.82
& \textbf{100.00} & 1.22 & 34.14 & \textbf{76.66} \\
\bottomrule
\end{tabular}
}
\end{table*}

\section{Benchmark}

\subsection{Experimental Setup}


To reduce shortcut cues unrelated to generation quality, we apply unified post-processing to all images. Specifically, all images are converted to PNG and resolution-aligned, with synthetic figures resized to match the corresponding real figure. For GPT-generated images, we remove the blank margins introduced during generation and retain only the central content region. For academic diagram-like figures, we further apply color quantization and color snapping to reduce nuisance variation: we first compress the color distribution using clustering-based quantization, then snap near-white and near-black pixels to canonical colors, merge remaining pixels toward dominant color clusters, and refine uniform color-block regions to improve boundary and intra-region consistency.
To prevent data leakage, we split the dataset at the paper level: all figures from the same paper, including overviews, illustrations, and experimental figures, are assigned to the same split. We follow a 10-fold cross-validation protocol, where each fold is divided into training, validation, and test sets with a ratio of 8:1:1. Unless otherwise noted, all results are averaged across the 10 folds.

\subsubsection{Evaluation Metrics.}
We follow existing works~\cite{wang2019cnngenerated,ojha2023towards,tan2024rethinking} in the field and report the average precision (AP) and classification accuracy (Acc) as the two main evaluation metrics. For accuracy, the classification threshold is fixed at 0.5 for each set to ensure a fair comparison.

\subsubsection{Evaluated Detectors}

To comprehensively assess the difficulty of AI-generated scientific figure detection, we benchmark a diverse set of representative AI-generated image detectors spanning spatial-domain, frequency-domain, gradient-based, structural-artifact, and foundation-model-based paradigms. Specifically, we include CNNSpot~\allowbreak\cite{wang2019cnngenerated}, PatchFor~\allowbreak\cite{chai2020makes}, UniFD~\allowbreak\cite{ojha2023towards}, LGrad~\allowbreak\cite{tan2023learning}, NPR~\allowbreak\cite{tan2024rethinking}, FreqNet~\allowbreak\cite{tan2024frequency}, FatFormer~\allowbreak\cite{liu2024forgery}, AIDE~\allowbreak\cite{yan2024sanity}, and Effort~\allowbreak\cite{yan2024effort}.

\subsection{Results and Analysis}

\subsubsection{Zero-shot Evaluation}

We first evaluate existing AI-generated image detectors in the zero-shot setting, where off-the-shelf models trained on prior AIGI datasets are directly applied to our benchmark without adaptation.

\textbf{Zero-shot transfer fails consistently across all detectors.}
As shown in Table~\ref{tab:zeroshot_main}, all methods suffer substantial degradation on our benchmark. Even the strongest model, LGrad, reaches only 53.68\% mAcc on the full set, while most other methods remain near chance level.
Existing detectors are heavily biased toward predicting scientific figures as real.
A striking pattern is the combination of very high $\mathrm{Acc}_{\text{real}}$ and extremely low $\mathrm{Acc}_{\text{fake}}$. For example, CNNSpot, PatchFor, FreqNet, FatFormer, and Effort all achieve near-perfect or perfect $\mathrm{Acc}_{\text{real}}$, but their $\mathrm{Acc}_{\text{fake}}$ is mostly below 3\%. This indicates that current detectors largely fail to recognize AI-generated scientific figures.

\textbf{The failure is systematic rather than category-specific.}
The same trend holds across illustrations, overviews, and experimental figures. For instance, Effort reaches 100.00\% $\mathrm{Acc}_{\text{real}}$ on all three categories, yet its fake accuracy drops to 0.00 on both illustrations and overviews. These results suggest a substantial domain gap between open-domain AIGI benchmarks and scientific figures.

\begin{table*}[t]
\centering
\caption{Cross-generator image classification under three training protocols. Models are evaluated on real, GPT-generated, and Nano Banana-generated 
scientific figures. $\mathrm{AvgAcc}$ denotes the unweighted mean of $\mathrm{Acc}_{real}$, 
$\mathrm{Acc}_{GPT}$, and $\mathrm{Acc}_{Banana}$, so that each subset contributes equally 
regardless of its size. All reported numbers are averaged over the 10 paper-level folds.}
\label{tab:cross_generator_all}
\resizebox{0.95\textwidth}{!}{
\begin{tabular}{lcccc|cccc|cccc}
\toprule
& \multicolumn{4}{c|}{Train on Banana}
& \multicolumn{4}{c|}{Train on GPT}
& \multicolumn{4}{c}{Train on Banana+GPT} \\
\cmidrule(lr){2-5} \cmidrule(lr){6-9} \cmidrule(lr){10-13}
Method
& Acc$_\text{real}$ & Acc$_\text{GPT}$ & Acc$_\text{Banana}$ & $\mathrm{AvgAcc}$
& Acc$_\text{real}$ & Acc$_\text{GPT}$ & Acc$_\text{Banana}$ & $\mathrm{AvgAcc}$
& Acc$_\text{real}$ & Acc$_\text{GPT}$ & Acc$_\text{Banana}$ & $\mathrm{AvgAcc}$ \\
\midrule
CNNSpot~\cite{wang2019cnngenerated}
& 98.67 & 40.04 & 78.46 & 72.39
& 97.93 & 77.95 & 11.11 & 62.33
& 94.84 & 78.75 & 73.85 & 82.48 \\
PatchFor~\cite{chai2020makes}
& 90.27 & 6.03  & 53.28 & 49.86
& 96.28 & \underline{98.92} & 8.38  & 67.86
& \underline{98.17} & \underline{95.14} & 11.58 & 68.30 \\
UniFD~\cite{ojha2023towards}
& 88.70 & 67.00 & 81.00 & \underline{78.90}
& 90.10 & 66.40 & 42.50 & 66.33
& 93.20 & 66.30 & 66.70 & 75.40 \\
LGrad~\cite{tan2023learning}
& 95.28 & 50.68 & 82.92 & 76.29
& 96.95 & 89.89 & 23.89 & 70.24
& 94.84 & 94.45 & 86.74 & 92.01 \\
NPR~\cite{tan2024rethinking}
& 65.80 & \textbf{87.69} & \textbf{98.93} & \textbf{84.14}
& 81.41 & 96.96 & \underline{51.21} & \underline{76.53}
& 93.75 & 93.43 & \textbf{94.69} & \underline{93.96} \\
FreqNet~\cite{tan2024frequency}
& 48.33 & \underline{73.83} & 87.98 & 70.05
& 87.92 & 82.73 & 26.05 & 65.57
& 81.45 & 83.83 & 83.84 & 83.04 \\
FatFormer~\cite{liu2024forgery}
& \underline{98.77} & 30.01 & 92.64 & 73.80
& \underline{99.00} & 92.79 & 7.86  & 66.55
& 97.59 & 90.45 & 79.62 & 89.22 \\
AIDE~\cite{yan2024sanity}
& 92.40 & 54.84 & 76.93 & 74.72
& 98.31 & 82.83 & 11.70 & 64.28
& 91.86 & 90.41 & 73.59 & 85.28 \\
Effort~\cite{yan2024effort}
& \textbf{99.72} & 28.40 & \underline{97.57} & 75.23
& \textbf{99.92} & \textbf{98.99} & \textbf{51.95} & \textbf{83.62}
& \textbf{98.30} & \textbf{95.81} & \underline{92.63} & \textbf{95.58} \\
\bottomrule
\end{tabular}
}
\end{table*}

\begin{table*}[h]
\centering
\caption{\textbf{Robustness under image degradation.} Classification accuracy (\%) on degraded test sets. Models are trained on clean \textbf{Banana+GPT} data and evaluated under JPEG/WebP compression, Gaussian blur, and Gaussian noise.}
\label{tab:degradation}
\resizebox{0.95\textwidth}{!}{
\begin{tabular}{l c cccc cccc ccc ccc}
\toprule
& & \multicolumn{4}{c}{JPEG Compression} & \multicolumn{4}{c}{WebP Compression} & \multicolumn{3}{c}{Gaussian Blur} & \multicolumn{3}{c}{Gaussian Noise} \\
\cmidrule(lr){3-6} \cmidrule(lr){7-10} \cmidrule(lr){11-13} \cmidrule(lr){14-16}
Method
& Clean
& q=95 & q=75 & q=50 & q=30
& q=95 & q=75 & q=50 & q=30
& $\sigma$=0.5 & $\sigma$=1.0 & $\sigma$=2.0
& $\sigma$=5 & $\sigma$=10 & $\sigma$=20 \\
\midrule
CNNSpot~\cite{wang2019cnngenerated}
& 82.50
& 80.78 & 79.56 & 78.17 & 75.74
& 77.80 & 77.07 & 75.41 & 72.78
& 80.99 & 80.92 & 83.45
& 56.74 & 55.28 & 56.12 \\
UniFD~\cite{ojha2023towards}
& 75.40
& 55.73 & 63.47 & 61.00 & 61.27
& 74.00 & 71.66 & 70.33 & 71.33
& 69.00 & 68.27 & 66.20
& 66.33 & 60.73 & 53.00 \\
NPR~\cite{tan2024rethinking}
& 93.96
& 87.38 & 81.51 & 78.76 & 75.94
& 83.29 & 78.22 & 73.04 & 66.86
& 91.65 & 89.54 & 79.13
& 69.05 & 65.45 & 50.49 \\
Effort~\cite{yan2024effort}    
& 95.58  
& 71.60 & 70.80 & 71.16 & 71.02 
& 69.69 & 69.33 & 69.08 & 68.25 
& 70.34 & 68.41 & 67.40   
& 66.75 & 66.75 & 66.71 \\
\bottomrule
\end{tabular}
}
\end{table*}

\label{sec:experiments}

\subsubsection{Cross-Generator Image Classification}

\begin{figure}[t]
\centering
\includegraphics[width=0.88\columnwidth]{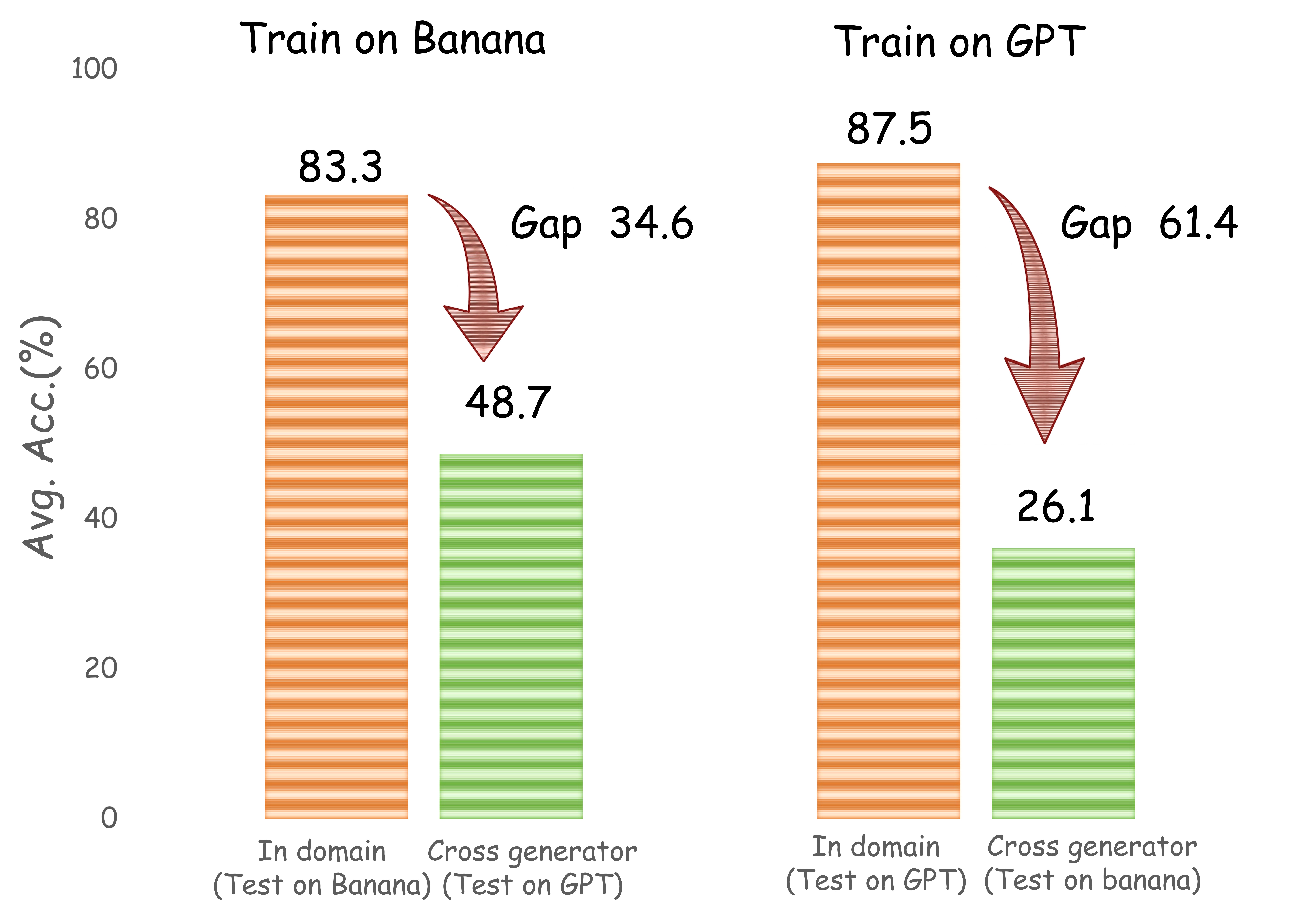}
\caption{\textbf{Cross-generator generalization gap.}
Averaged across detectors, models trained on one generator perform much better on the seen generator than on the unseen one. This large in-domain vs.\ cross-generator gap indicates strong generator-specific overfitting and a clear domain gap between Banana- and GPT-generated scientific figures.}
\vspace{-1em}
\label{fig:cross_generator_gap}
\end{figure}

Single-generator training leads to strong generator overfitting.
Tables~\ref{tab:cross_generator_all} show that detectors trained on one generator often fail to transfer to the other. For example, when trained on Banana, Effort drops from 97.57\% on Banana to only 28.40\% on GPT; This indicates that many existing detectors still rely on generator-specific cues rather than generator-invariant principles, suggesting that generalization to future unseen generators remains a major challenge.
Figure~\ref{fig:cross_generator_gap} summarizes the cross-generator gap by averaging over all detectors. Training on Banana gives 83.3\% on Banana but only 48.7\% on GPT, while training on GPT gives 87.5\% on GPT but only 26.1\% on Banana. This large drop confirms strong generator-specific overfitting and suggests a clear domain gap between Banana- and GPT-generated scientific figures.

\textbf{Joint training improves in-domain robustness, but the problem is far from saturated.}
Training on both Banana and GPT consistently improves performance for most methods (Table~\ref{tab:cross_generator_all}). In particular, Effort achieves the best overall average accuracy of 95.58\%, followed by NPR  and LGrad, showing that exposure to multiple generators is important for this benchmark. However, these gains should be interpreted as improved performance on \emph{seen} generators rather than solved generalization: the strong failures in the single-generator setting indicate that detectors may still struggle when confronted with new generators outside the training set.

The remaining gap under joint training suggests a non-trivial domain gap between Banana and GPT.
Even after jointly training on both sources, performance does not fully saturate, and robustness remains model-dependent. While Effort, NPR, and LGrad become more balanced across real, GPT, and Banana subsets, other methods remain brittle. PatchFor is a representative failure case: despite 98.17\% accuracy on real images and 95.14\% on GPT, it achieves only 11.58\% on Banana.  These results show that Banana and GPT generated scientific figures are not a single homogeneous fake image distribution. High accuracy on seen generators therefore does not imply robust cross-generator generalization.

\subsubsection{Degraded Image Classification}

We evaluate robustness by testing detectors trained on clean (Banana+GPT) data under common post-processing corruptions, including JPEG compression, WebP compression, Gaussian blur, and Gaussian noise. Table~\ref{tab:degradation} reports classification accuracy on degraded test sets. This setting reflects practical deployment scenarios in which scientific figures may undergo re-saving, format conversion, document rendering, or screenshot-based redistribution.
\textbf{Image degradation remains a major failure mode.}
Most detectors suffer clear performance drops once the test images are corrupted. For example, NPR drops from 93.96\% on clean images to 75.94\% under JPEG compression at q=30, and further to 50.49\% under Gaussian noise with $\sigma=20$. UniFD is even more sensitive to compression, falling from 71.93\% on clean images to 55.73\% at JPEG q=95.

\textbf{Strong clean performance does not imply robustness.}
Effort achieves the best clean accuracy, but its performance drops to around 68--72\% under JPEG and WebP compression. CNNSpot is relatively stable under compression and blur, but degrades substantially under Gaussian noise. NPR remains strong under mild blur and compression, yet fails under severe noise. Overall, these results show that robustness to realistic post-processing remains unsolved, even for detectors that perform well on clean data.

\section{Conclusion}

We introduced the first benchmark for AI-generated scientific figure detection, targeting a new and increasingly practical threat model enabled by modern multimodal generators. To support this setting, we developed a scalable agent-based data construction pipeline that builds high-quality real--synthetic figure pairs from licensed papers. Our experiments show that existing detectors fail in the zero-shot setting, generalize poorly across generators, and remain fragile under common image degradations. We hope this benchmark can serve as a foundation for future research on more robust and generalizable scientific-figure forensics.


\bibliographystyle{ACM-Reference-Format}
\bibliography{sample-base}




\end{document}